\documentclass[sigconf]{acmart}

\def\eqref#1{equation~\ref{#1}}

\def\1{\bm{1}}

\DeclareMathAlphabet{\mathsfit}{\encodingdefault}{\sfdefault}{m}{sl}
\SetMathAlphabet{\mathsfit}{bold}{\encodingdefault}{\sfdefault}{bx}{n}

\usepackage{hyperref}
\usepackage{url}

\usepackage[utf8]{inputenc} %
\usepackage{booktabs}       %
\usepackage{nicefrac}       %
\usepackage{microtype}      %
\usepackage{multirow}
\usepackage{graphicx}
\usepackage[normalem]{ulem} %
\usepackage{enumitem}
\usepackage{subcaption}
\usepackage{wrapfig}
\usepackage{colortbl} %
\definecolor{lightgray}{rgb}{0.9,0.9,0.9}

\renewcommand{\thefootnote}{\fnsymbol{footnote}}
\newcommand{\fix}{\color{black}}

\AtBeginDocument{%
  }

\setcopyright{none}
\acmConference{Submission \#15}{KDD Datasets and Benchmarks}{2025}
\acmISBN{}
\acmDOI{}

\renewcommand\footnotetextcopyrightpermission[1]{}

\begin{document}

\title{CityBench: Evaluating the Capabilities of Large Language Models for Urban Tasks}

\author{Jie Feng*}
\affiliation{%
 \institution{Department of Electronic\\Engineering, BNRist,\\Tsinghua University}
 \country{Beijing, China}
}
\email{fengjie@tsinghua.edu.cn}

\author{Jun Zhang*}
\affiliation{%
 \institution{Department of Electronic\\Engineering, BNRist,\\Tsinghua University}
 \country{Beijing, China}
}
\email{zhangjun990222@gmail.com}

\author{Tianhui Liu*}
\affiliation{%
 \institution{School of Electronic and Information Engineering,\\Beijing Jiaotong University}
 \country{Beijing, China}
}
\email{21211125@bjtu.edu.cn}

\author{Xin Zhang$\dagger$}
\affiliation{%
 \institution{Shenzhen International \\Graduate School,\\Tsinghua University}
 \country{Shenzhen, China}
}
\email{zhangxin4087@163.com}

\author{Tianjian Ouyang$\dagger$}
\affiliation{%
 \institution{Department of Electronic\\Engineering, BNRist,\\Tsinghua University}
 \country{Beijing, China}
}
\email{oytj22@mails.tsinghua.edu.cn}

\author{Junbo Yan$\dagger$}
\affiliation{%
 \institution{Department of Electronic\\Engineering, BNRist,\\Tsinghua University}
 \country{Beijing, China}
}
\email{yanjb20@mails.tsinghua.edu.cn}

\author{Yuwei Du$\dagger$}
\affiliation{%
 \institution{Department of Electronic\\Engineering, BNRist,\\Tsinghua University}
 \country{Beijing, China}
}
\email{duyw23@mails.tsinghua.edu.cn}

\author{Siqi Guo$\dagger$}
\affiliation{%
 \institution{Department of Electronic\\Engineering,\\Tsinghua University}
 \country{Beijing, China}
}
\email{guosq21@mails.tsinghua.edu.cn}

\author{Yong Li$\ddagger$}
\affiliation{%
 \institution{Department of Electronic Engineering, BNRist,\\Tsinghua University}
 \country{Beijing, China}
}
\email{liyong07@tsinghua.edu.cn}

\renewcommand{\shortauthors}{Jie Feng et al.}

\begin{abstract}
As large language models (LLMs) continue to advance and gain widespread use, establishing systematic and reliable evaluation methodologies for LLMs and vision-language models (VLMs) has become essential to ensure their real-world effectiveness and reliability. There have been some early explorations about the usability of LLMs for limited urban tasks, but a systematic and scalable evaluation benchmark is still lacking. The challenge in constructing a systematic evaluation benchmark for urban research lies in the diversity of urban data, the complexity of application scenarios and the highly dynamic nature of the urban environment. In this paper, we design \textit{CityBench}, an interactive simulator based evaluation platform, as the first systematic benchmark for evaluating the capabilities of LLMs for diverse tasks in urban research. First, we build \textit{CityData} to integrate the diverse urban data and \textit{CitySimu} to simulate fine-grained urban dynamics. Based on \textit{CityData} and \textit{CitySimu}, we design 8 representative urban tasks in 2 categories of perception-understanding and decision-making as the \textit{CityBench}. With extensive results from 30 well-known LLMs and VLMs in 13 cities around the world, we find that advanced LLMs and VLMs can achieve competitive performance in diverse urban tasks requiring commonsense and semantic understanding abilities, e.g., understanding the human dynamics and semantic inference of urban images. Meanwhile, they fail to solve the challenging urban tasks requiring professional knowledge and high-level numerical abilities, e.g., geospatial prediction and traffic control task. These findings provide critical insights for the effective utilization and further development of LLMs to advance urban-related tasks and research in the future. The associated data and code are publicly available at \url{https://github.com/tsinghua-fib-lab/CityBench}.
\end{abstract}

\maketitle

\footnotetext[1]{These authors contributed equally.}
\footnotetext[2]{These authors contributed equally.}
\footnotetext[3]{Corresponding author, email: liyong07@tsinghua.edu.cn}

\renewcommand{\thefootnote}{\arabic{footnote}}
\setcounter{footnote}{0}

\section{Introduction} \label{sec:intro}
Recent years, large language models (LLMs) with extensive commonsense and reasoning capabilities have achieved excellent results in various fields~\citep{achiam2023gpt,touvron2023llama}, including programming~\citep{hong2023metagpt}, mathematics~\citep{wei2022emergent}, visual intelligence~\citep{liu2024visual} and commonsense reasoning~\citep{suzgun2022challenging, mialon2023gaia}. Furthermore, powerful LLMs enable many unimaginable research endeavors to become feasible, e.g., agent~\citep{wang2024survey} and embodied intelligence~\citep{reed2022generalist, zha2025enable}. These researchers postulate that LLMs, by acquiring extensive world knowledge and commonsense, hold the key to unlocking promising outcomes in these challenging applications. Many works~\citep{achiam2023gpt, gurnee2023language} have demonstrated that LLMs can be regarded as `world models' of our life and they are skilled at solving a wide variety of tasks across multiple fields, while other works~\citep{xiang2024language,yang2023learning,wang2024can} indicate that LLMs lack an comprehensive understanding of the real physical world and fail to handle many real-life problems. However, these research efforts have primarily focused on the indoor environment~\citep{puig2018virtualhome}, while neglecting the outdoor environment, specifically the broader urban environment~\citep{batty2012smart,zheng2014urban}.

Various works have explored the potential of LLMs in modeling urban space and solving urban tasks. For example, researchers evaluate the potential of LLMs on remote sensing understanding tasks~\citep{kuckreja2023geochat} and urban visual tasks~\citep{yan2024urbanclip}. Gurnee et al.~\citep{gurnee2023language} evaluate whether LLMs acquire the spatial knowledge of the world, such as cities and coordinates. Manvi et al.~\citep{manvi2023geollm, manvi2024large} try to extract the geospatial knowledge in LLMs to conduct geospatial indicator prediction tasks~\citep{mai2023opportunities}. Besides, researchers also explore how to apply LLMs into the realistic urban applications, e.g., traffic control~\citep{lai2023large}, mobility prediction~\citep{wang2023would, feng2025agentmove}, behavior modeling~\citep{gong2024population}, visual language navigation~\citep{schumann2024velma} and so on. However, on the one hand, these existing works primarily focus on evaluating the static spatial knowledge of LLMs without considering the environment dynamics and interactivity. On the other hand, most of them only focus on one type of task and one modality of data in the urban space, using small dataset that are not scalable globally. Although there are some existing simulators for urban space such as game simulators~\citep{skyline} and traffic simulators~\citep{SUMO2018}, they cannot be directly applied to support the evaluation and significant amount of adaptation work is required. None of them can support the systematic evaluation of LLMs' capabilities for diverse tasks in urban research, ranging from understanding and reasoning to decision-making tasks.

\begin{figure*}
    \centering
    \includegraphics[width=1\textwidth]{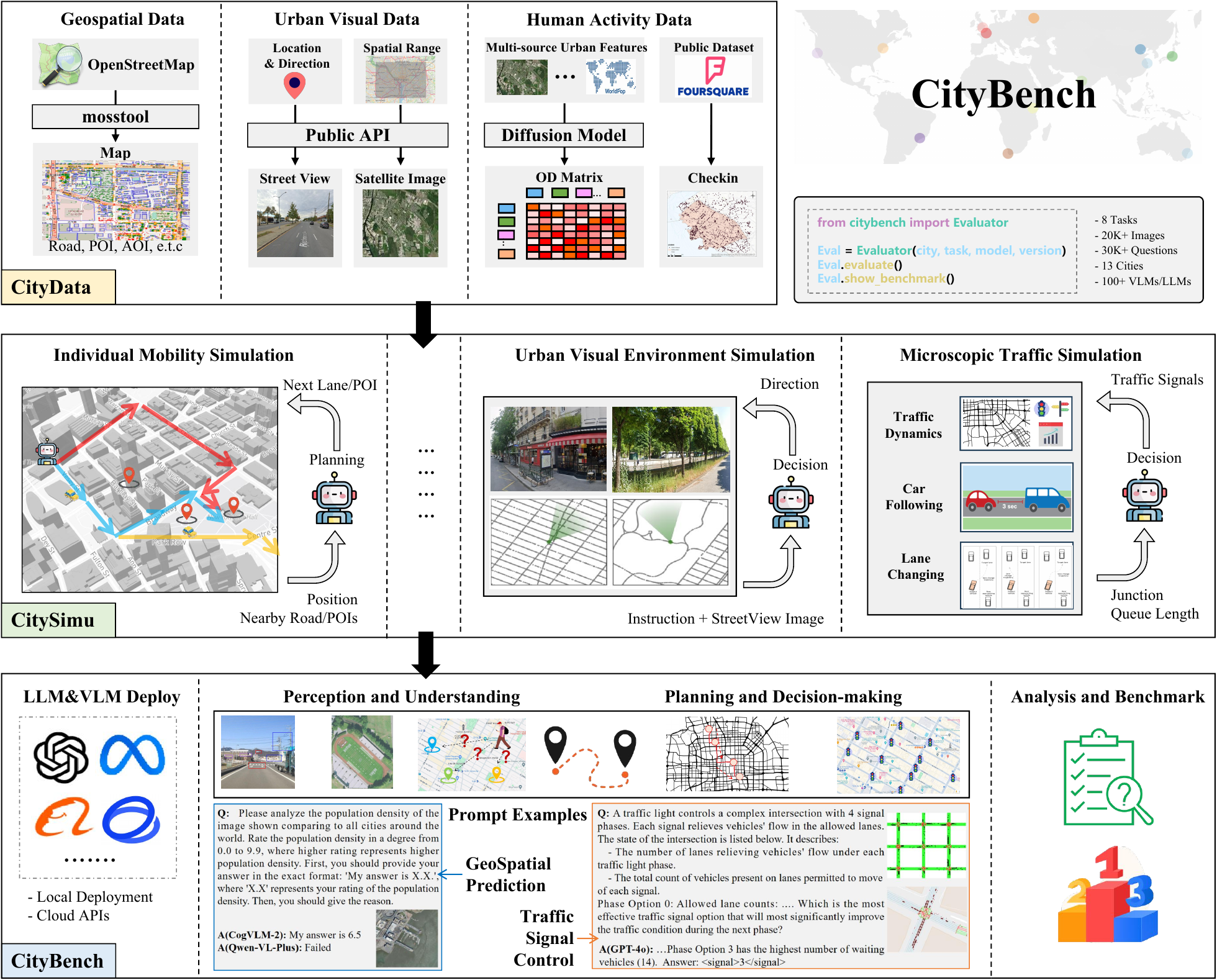}
    \caption{The framework of \textbf{\textit{CityBench}}, which consists of a data collector~\textit{CityData}, an activity simulator~\textit{CitySimu} and 8 diverse urban tasks with different modalities. The evaluation data in the benchmark is collected from 13 cities around the world.}
    \label{fig:framework}
\end{figure*}

In this paper, we propose \textbf{\textit{CityBench}}, a comprehensive evaluation platform for assessing the capabilities of LLMs to solve the diverse urban tasks. It covers multiple modalities, supports interactive simulations, and includes data from 13 cities around the world. \textit{CityBench} consists of three modules: a data module~\textit{CityData} for collecting and processing diverse urban data, a simulation module \textit{CitySimu} for simulating fine-grained urban dynamics, a evaluation module \textit{CityBench} for the final evaluation of LLMs and VLMs. In \textit{CityData}, we first collect three kinds of open urban data: geospaital data from Open Street Map, urban visual data from the Google map and ArcGIS, and human activity data from Foursquare and other websites. Then, we build an efficient simulation engine \textit{CitySimu} to simulate fine-grained urban dynamics and develop various interfaces for controlling the urban dynamics and sensing the urban environments. Furthermore, based on \textit{CitySimu}, we design a comprehensive benchmark to evaluate the capability of LLMs and VLMs, covering core research problems from various urban research fields. The benchmark comprises two levels of tasks: perception\&understanding tasks and decision-making tasks. In perception\&understanding tasks, based on the integrated multi-source data from \textit{CitySimu}, we introduce street view and satellite image understanding and urban space understanding tasks to evaluate the urban geospatial knowledge of LLMs and VLMs. In decision-making tasks, we apply LLMs and VLMs to interact with \textit{CitySimu} to complete the urban exploration, visual navigation, mobility prediction and traffic signal control task,  which require the comprehensive ability of them. In summary, our contribution are as follows,

\begin{itemize}[leftmargin=1.5em,itemsep=1pt,parsep=0.2em,topsep=0.5em,partopsep=0.0em]
    \item We develop \textit{CityData} and \textit{CitySimu}, an urban data collector and processor designed to support diverse urban tasks and applications, as well as an efficient urban simulator for generating find-grained urban dynamics. They provide ease-to-use APIs for controlling urban dynamics and sensing urban environments.
    \item We propose \textit{CityBench}, a comprehensive evaluation benchmark for evaluating the capability of LLMs and VLMs for urban tasks, which includes 4 geospatial understanding tasks and 4 interactive urban decision-making tasks in 13 cities around the world.
    \item Extensive experiments on \textit{CityBench} with 30 well-known open source and proprietary  LLMs and VLMs demonstrate the effectiveness of \textit{CityBench} as evaluation benchmark and also discuss the potential and limitation of applying LLMs and VLMs in urban tasks, ranging from understanding and reasoning to decision-making task.
\end{itemize}

\begin{figure*}
    \centering
    \includegraphics[width=1\textwidth]{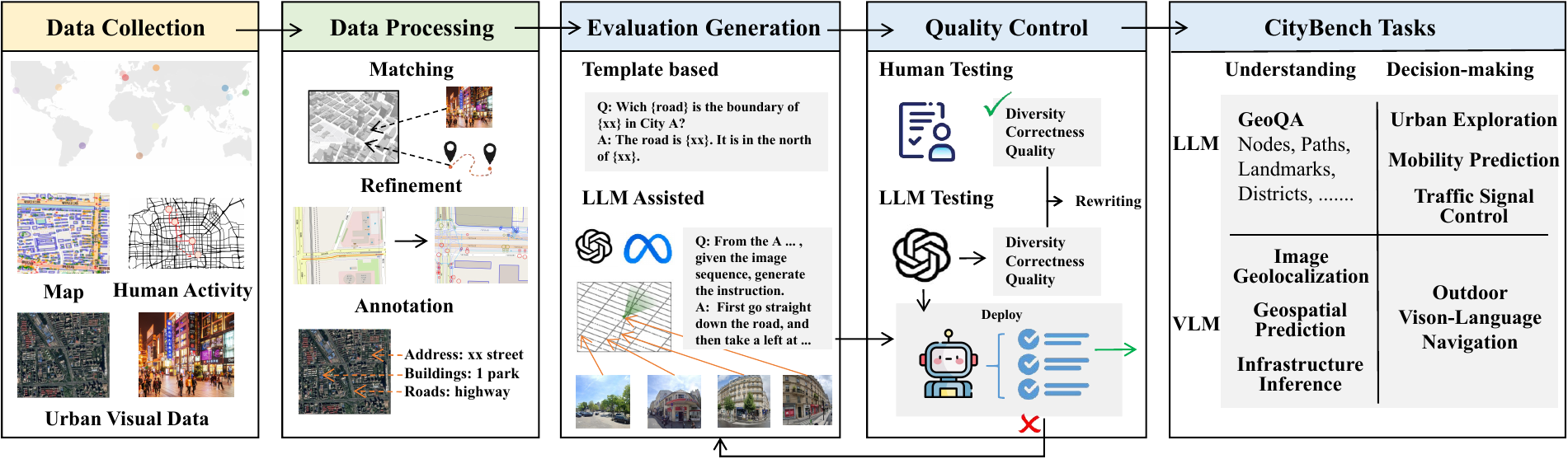}
    \caption{The pipeline of building benchmark, including data collection stage, data integration stage, evaluation generation stage and quality control stage.}
    \label{fig:benchamark}
\end{figure*}

\section{Methods}

As presented in Figure~\ref{fig:framework}, \textbf{\textit{CityBench}} is a simulator based evaluation platform with three core components: \textit{CityData} for collecting and processing diverse urban data, \textit{CitySimu} for simulating human dynamics and providing an interactive simulation environment, and \textit{CityBench} for model evaluation on 8 representative urban tasks with different modalities.

\subsection{CityData}\label{sec:citydata}

In the section, we introduce the multi-source urban dataset \textit{CityData} collected to support multi-modal urban tasks. 
To present a complete picture of the city's geospatial structure, semantic features, and human activities, \textit{CityData} integrates the following globally available data from multiple sources~\citep{liu2023urbankg}. The python package for data collection and access is open-source\footnote{\url{https://github.com/tsinghua-fib-lab/pycitydata}}.

\textbf{Geospatial Data}
Geospatial data, represented by maps, is the most fundamental data for describing the urban structure including road networks, points of interest (POIs), areas of interest (AOIs), etc.
OpenStreetMap (OSM)~\footnote{https://www.openstreetmap.org/} is most widely used open source map data.
However, the raw data provided by OSM cannot support the simulation of urban dynamics directly because the relationship between different elements is incomplete such as the connection between buildings and roads.
Therefore, we provide a globally available rule-based map building tool~ \footnote{\url{https://github.com/tsinghua-fib-lab/mosstool}} within \textit{CityData} that reconstructs lanes, lane topology, and building-lane connections based on the raw OSM data. The reconstructed map is used as the geospatial base and simulation input in \textit{CitySimu}.

\textbf{Urban Visual Data}
Street view data and satellite images are two types of globally available urban data that contains rich semantic information, which represents the visual of human.
Therefore, \textit{CityData} also integrates the two types of data, the former obtained via Google Maps API and Baidu Maps API, and the latter using the Esri World Imagery as data source.
In \textit{CityData}, street view data is accessed through spatial location and facing direction, and satellite images are acquired through spatial ranges.

\textbf{Human Activities Data} \label{sec:human}
\fix We use the global Foursquare-checkin~\citep{yang2016participatory} data and a synthetic global origin-destination data (OD data)~\footnote{\url{https://github.com/tsinghua-fib-lab/generate-od-pubtools}} as the proxy of human activities to enable the fine-grained human movement simulation.
The Foursquare-checkin dataset~\citep{yang2016participatory} is a long-term user check-in dataset collected from Foursquare~\footnote{\url{https://foursquare.com/}} in approximately 400 cities worldwide. It has been widely used in the community over the past ten years \citep{chen2024deep}. \color{black}
Origin-destination data is generated by a diffusion model with population from Worldpop~\footnote{\url{https://www.worldpop.org/}} and satellite image from Esri World Imagery as input.
While all the user information are anonymized, we follow the license from Foursquare-checkin~\citep{yang2016participatory} to protect the public privacy.

\begin{figure*}
    \centering
    \includegraphics[width=1\textwidth]{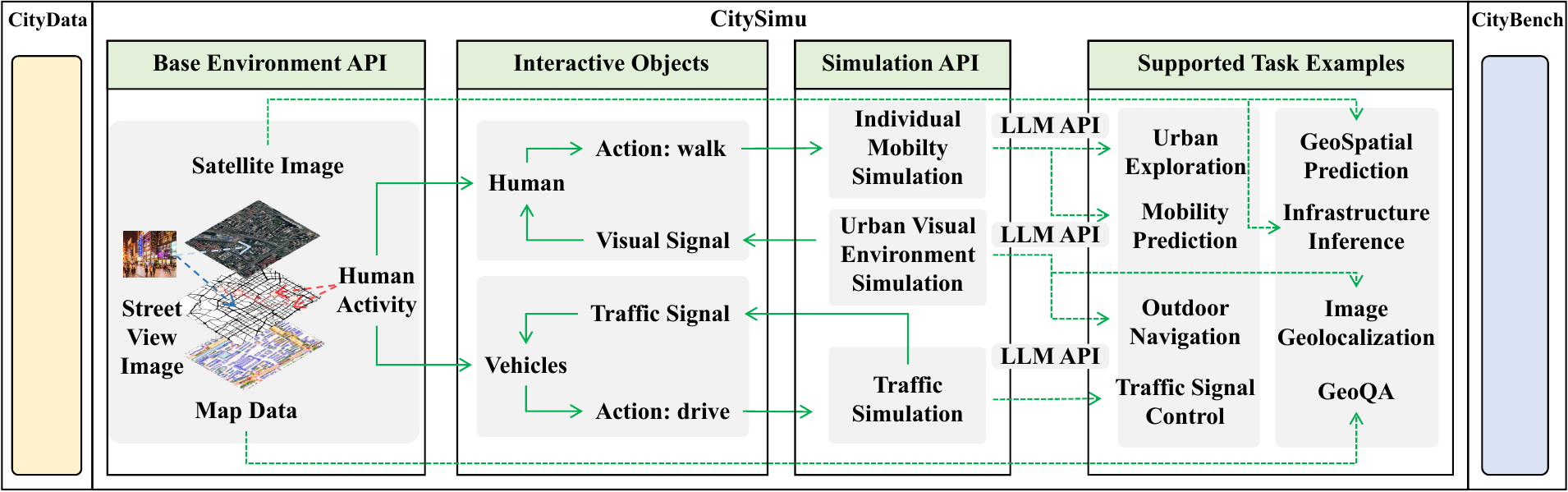}
    \caption{The simulation framework of \textit{CitySimu}, including base environment APIs, interactive objects, simulation APIs and language APIs. Besides, supported task examples also present the relation between simulation APIs and evaluation tasks.}
    \label{fig:citysim}
\end{figure*}

\subsection{CitySimu}\label{sec:citysim}

Building on \textit{CityData}, \textit{CitySimu} simulates the urban dynamics and provide diverse easy-to-use APIs for the interactive operation. As shown in Figure~\ref{fig:citysim}, \textit{CitySimu} contains the base environment APIs for obtaining the static information of environment, three simulation APIs for human and vehicle behavior simulations, language APIs to enable the interaction between \textit{CitySimu} and LLMs.

\textbf{Individual Mobility Simulation}
Based on the geospatial data, the individual mobility simulation constructs a simulator that can simulate an agent moving and exploring within the city.
Agents can obtain the POIs and roads around them through API provided by \textit{CityData}, and thus plan and decide the next lane or POI to travel in to update their locations. For the mobility prediction task in the city scale, the available actions are defined as the POIs around the city. For the urban exploration task in the local street scale, the available actions are defined as the nearby lanes.

\textbf{Urban Visual Environment Simulation}
To further support the study of urban visual intelligence~\citep{fan2023urban}, we follow ~\citep{chen2019touchdown, mirowski2018learning} to construct a urban visual environment simulation with real street view images and map data. In the environment, agent can access the panoramic images of its location via APIs and then select the available actions to move along the road to arrive the destination. In the outdoor visual-language instruction navigation task, given the human-like instruction, agent can observe the panoramic images of its location, extract key elements from them and then decide one direction to go. This can be saw as an extension of individual mobility simulation with visual input. 

\textbf{Traffic Simulation}
In the former two simulations, we only simulate the individual actions without the interaction with others. Here, we introduce microscopic traffic simulation to model the interaction behaviors between vehicles and provide a traffic control environment. 
The simulator takes the geospatial data reconstructed from OSM within \textit{CityData} and the travel demand described by the synthetic global OD data as inputs. It simulates the vehicle behaviors through realistic driving simulation models including the intelligent driver model (IDM)~\citep{treiber2000congested} as the car-following model and the randomized MOBIL model~\citep{kesting2007general,feng2021intelligent} as the lane-change model to obtain the dynamics of all vehicles in the city at each second. The simulator also provides a series of sensing and control APIs. Through the sensing APIs, LLMs can obtain data about urban dynamics such as junction queue length, vehicle speed, and road average speed. Through the control APIs, LLMs can intervene in the city's operation, such as modifying traffic signal lights, modifying the speed limit of the road, etc.

\subsection{CityBench} \label{sec:citybench}

Based on \textit{CityData} and \textit{CitySimu}, we design a multi-modal urban evaluation benchmark \textit{CityBench} to evaluate the capability of LLMs and VLMs. In the following section, we first summarize the whole pipeline and then give introduction to each task. 

\subsubsection{Pipeline}

Figure~\ref{fig:benchamark} describe the procedure of building evaluation benchmark. As introduced before, \textit{CityData} works in the data collection and data processing stage and \textit{CitySimu} works in the data processing stage. We focus on introducing the evaluation generation stage and quality control stage as follows.

In the evaluation generation stage, we use template based methods and LLMs/VLMs based methods to generate the evaluation questions. For example, for the image geolocalization task, the groundth location is already known when collecting, thus we directly design template based question to convert the image geolocalization task into question answer pair. As for the outdoor navigation task, we employ VLM to act as human annotation experts to annotate the data to generate the navigation instruction with additional inputs. In \textit{CityBench}, instructions for urban exploration task and outdoor navigation task are generated by LLM assisted methods. Instructions for other tasks are generated from template based methods.

Due to the hand-craft designs and potential issues of LLMs, we apply a quality control stage to filter and rewrite the generated questions to obtain a high quality evaluation questions. For questions generated from template based methods, we use LLM as data quality expert to filter the low-quality data and use LLM as data rewritter to rewrite the questions with diverse formats and expressions. For questions generated from the LLMs/VLMs based methods, we use LLM/VLM as the agent with additional information to execute the task to verify the quality of generated instructions. If the generated questions are filtered too much, we will return to the evaluation generation stage to generate new questions again. Finally, authors of this paper also participate in the quality control stage to filter and rewrite the generated data to ensure the quality of whole benchmark.

\begin{figure}
  \centering
    \includegraphics[width=0.43\textwidth]{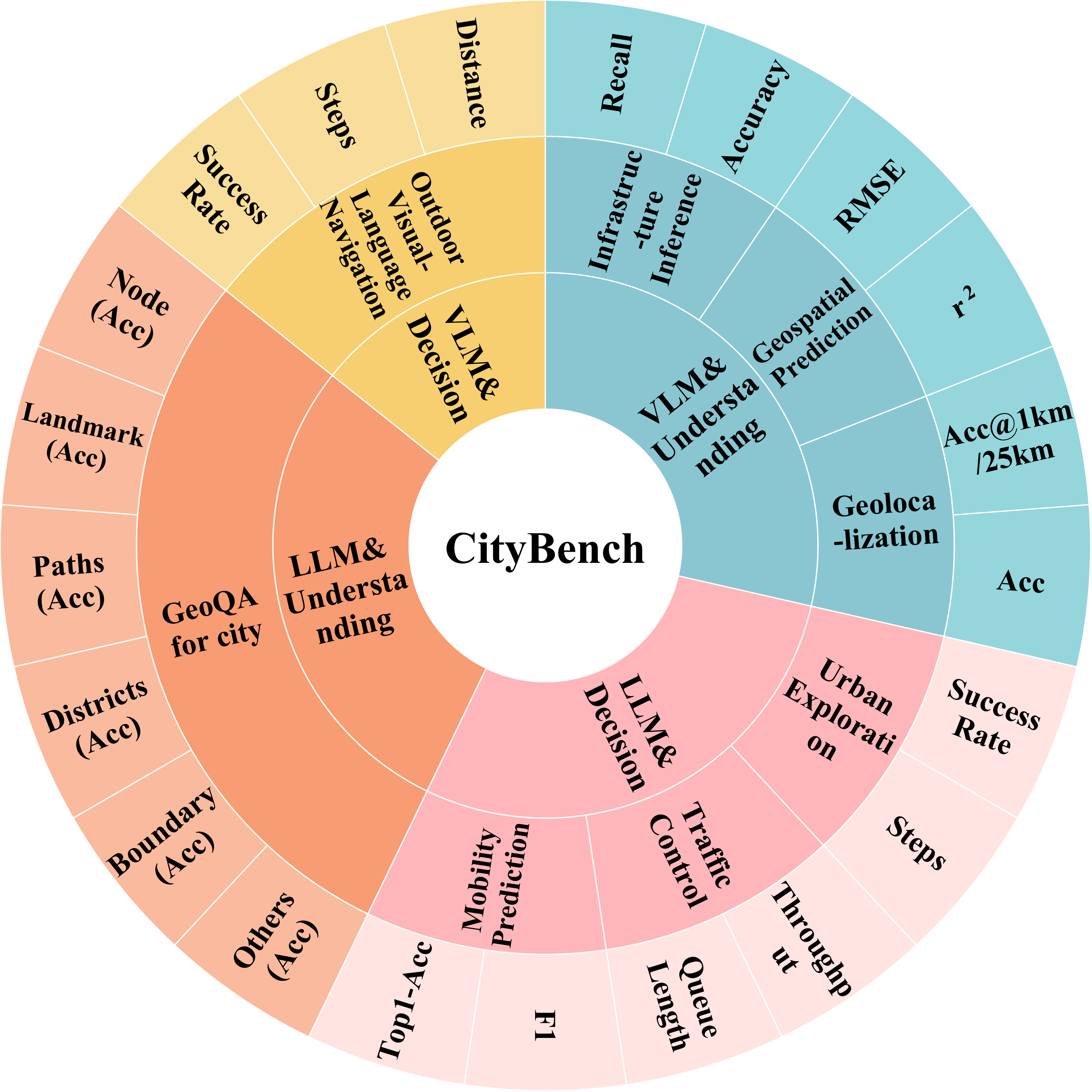}
    \caption{8 tasks in \textit{CityBench} with their metrics.}
    \label{fig:category}
\end{figure}

After the above stages, we produce the evaluation benchmark with 8 urban tasks. Their relations are presented in Figure~\ref{fig:category}. Details of each task are introduced as follows.

\subsubsection{Perception and Understanding Task}

The first task is the street view image geolocalization task from the urban visual intelligence~\citep{fan2023urban}. Following are social indicator prediction and infrastructure inference tasks from remote sensing field. Finally, we adapt GeoQA~\citep{mai2021geographic, feng2024citygpt} task into urban environment.

\textbf{Image Geolocalization}
Image geolocalization task is to predict the precise location of image based on its context. Street view image is regarded as the recording of urban appearance and play an important role in understanding the urban environment and dynamics~\citep{fan2023urban}. Thus, we query VLMs with street view image and require them to directly generate the location of image. A good VLM should recognize the important objects from the street image and mapping them into the potential locations. Following ~\citep{haas2023pigeon}, we define two subtasks for this task: city name inference and precise latitude and longitude inference.

\textbf{Geospatial Prediction}
Geospatial predictions are important for understanding the global sustainable development especially for developing countries, e.g., poverty estimation~\citep{jean2016combining} and population density estimation~\citep{tatem2017worldpop}. One of the most widely used solutions is using satellite images with machine learning methods to predict these socioeconomic indicators. In the benchmark, following setting from~\citep{manvi2023geollm}, we query VLMs with a satellite image as context to predict the population density of it. We use population from Worldpop~\citep{tatem2017worldpop} as the groundtruth.

\textbf{Infrastructure Inference}
Besides, we also introduce the infrastructure inference task which means to recognize the urban infrastructures from the satellite images. 
This task require the ability of scene understand and object segmentation of urban environment. 
The groundtruth of this task is extracted from the OSM by matching predefined infrastructure key words within a fixed spatial range. Given the satellite image and a list of all kinds of infrastructures, VLM is required to generate the infrastructure names appeared in the image. 
Here, we pay attention to the following infrastructures: Airport, Harbor, Stadium, Bridge, Roundabout and Train Station.

\textbf{GeoQA for City Elements}
Beyond understanding the urban space from the visual perspective, we introduce geographic question answer(GeoQA)~\citep{mai2021geographic} to test whether LLMs comprehends the fundamental elements~\citep{lynch1964image} in a city from the concept view, such as road and landmarks. 
For example, we directly ask LLM about the relation between different roads in a city. 
\fix Following~\citep{lynch1964image, feng2024citygpt}, we classify the spatial elements into six groups and design problems for each group. \color{black}
These six groups are node, path, landmark, boundary, districts and others.

\subsubsection{Planning and Decision Making Task}

Different from the static evaluation introduced in the last section, we design four interactive decision making tasks to evaluate the capabilities of LLMs in dynamic and partial observed environments which are more challenging and realistic. With the interaction with the \textit{CitySimu} and dynamic human activities, LLMs need to understand the important mechanisms and regularity in the urban environments to complete the decision-making tasks. 

\textbf{Mobility Prediction}
As one of the fundamental task for understanding the human behaviors and urban dynamics, mobility prediction task is to predict the next location of user in the next time window with given the past mobility trajectory. Here we use the the global Foursquare checkin data to support the mobility prediction in the simulator. We follow~\citep{wang2023would} to conduct the mobility prediction task via LLMs. 

\textbf{Outdoor Navigation}
Outdoor navigation task is widely used in neurocognitive science~\citep{epstein2017cognitive} as the important benchmark for evaluating the spatial cognition of human and models. As one of the most widely-used settings in outdoor navigation task, vision-language navigation task~\citep{schumann2024velma, yang2024v} requires the model to follow the human-annotated language instruction to arrive to the destination with the nearby street view images as additional input. This task requires the VLMs to acquire the ability of urban visual scene understanding, language understanding and decision-making.

\textbf{Urban Exploration}
Here, we define a text based street exploration task to evaluate the zero-shot navigation capability of LLMs in a new city without visual input and instructions. Different from the visual language navigation which require model to follow the language instruction and understand the scene via street view image, our urban exploration task require model to explore the region via the local information (e.g. accessed road names) provided by the simulator during action and its intrinsic knowledge of the whole urban space in the city.

\textbf{Traffic Signal Control}
Traffic signal control task is one of the widely studied realistic urban decision making task in recent years~\citep{wei2019survey}. It is challenging for existing methods due to the dynamic traffics and the generalization issues. It is to generate the future traffic signal schedule by considering the current traffic states and the future traffics. Lai et al.~\citep{lai2023large} propose LLMLight to employ LLM as decision-making agent for traffic signal control problem and demonstrate the generalization of LLMs. Following this work, we evaluate the potential of LLMs as agents for multiple-intersections traffic signal control.

\section{Benchmark and Experiments} \label{sec:exp}
\begin{table}
\centering
\caption{Detailed statistics of multi-source data for 13 global cities utilized in \textit{CityBench}.
}

\label{table:data}
\resizebox{0.48\textwidth}{!}{
\setlength{\tabcolsep}{1mm}
\begin{tabular}{lcccccc} 
\toprule
 \multicolumn{1}{c}{\multirow{2}{*}{\textbf{Cities}}} & \multicolumn{2}{c}{\textbf{Visual Data }} & \multicolumn{2}{c}{\textbf{GeoSpatial Data}} & \multicolumn{2}{c}{\textbf{Human Activity Data }} \\
 \multicolumn{1}{c}{} & \begin{tabular}[c]{@{}c@{}}\textbf{Satellite }\\\textbf{Image}\end{tabular} & \begin{tabular}[c]{@{}c@{}}\textbf{StreetView}\\\textbf{Image}\end{tabular}& \textbf{Roads} & \textbf{PoI/AoIs} & \begin{tabular}[c]{@{}c@{}}\textbf{OD flow }\\\textbf{(\textgreater{}10)}\end{tabular} & \textbf{Checkins} \\ 
\midrule
 \textbf{Beijing} & 1764 & 7482& 17043 & 276090 & 1905025 & 21015 \\
  \textbf{Shanghai} & 5925 & 4170& 33321 & 57731 & 845188 & 33129 \\
  \textbf{Mumbai} & 638 & 6025& 6296 & 60245 & 309147 & 31521 \\
  \textbf{Tokyo} & 1120 & 5514& 33174 & 1146094 & 969865 & 1044809 \\ 
\midrule
 \textbf{London} & 1710 & 4148& 14418 & 83892 & 1401404 & 173268 \\
  \textbf{Paris} & 238 & 6044& 4443 & 21950 & 28362 & 85679 \\
  \textbf{Moscow} & 1558 & 5761& 9850 & 28289 & 979064 & 836313 \\ 
\midrule
 \textbf{NewYork} & 320 & 3934& 5414 & 349348 & 71705 & 390934 \\
  \textbf{SanFrancisco} & 345 & 4473& 4171 & 73777 & 61367 & 100249 \\
  \textbf{SaoPaulo} & 1332 & 5184& 28714 & 1681735 & ~311830 & 808754 \\ 
\midrule
 \textbf{Nairobi} & 336 & 5987& 2972 & ~264101 & 135332 & 25727 \\
  \textbf{CapeTown} & 896 & 5175& 5947 & 151711 & 525578 & 11591 \\ 
\midrule
 \textbf{Sydney} & 1935 & 5087& 21390 & 141997 & 438763 & 54170 \\
\bottomrule
\end{tabular}
}
\end{table}

\subsection{Settings} \label{sec:api}

\textbf{Model Deployment} To facilitate usage of \textit{CityBench}, we have implemented local deployment support for the majority of LLMs and VLMs using VLMEvalKit~\citep{duan2024vlmevalkit} and vLLM~\citep{kwon2023efficient}. Additionally, we also support evaluation through the APIs of proprietary models, e.g., OpenAI and open-source models, e.g. DeepInfra~\footnote{\url{https://deepinfra.com/}} and Siliconflow~\footnote{https://siliconflow.cn/}.

\textbf{Baselines} We select well-known LLMs and VLMs as baselines. For VLMs, we select LLaVa-NeXT~\citep{liu2024visual}, CogVLM-v2~\citep{wang2023cogvlm},  MiniCPM-LLama3-V-2.5~\citep{MiniCPM-v}, Qwen-VL-plus and GPT4o. For LLMs, we select LLama3-8B, LLama3-70B, Mistral-7B-v0.2~\citep{jiang2023mistral}, Mixtral-8x22B-v0.1~\citep{jiang2024mixtral}, DeepSeekv2~\citep{Shao2024DeepSeekV2AS}, GPT3.5, and GPT4~\citep{achiam2023gpt}. We also select representative baselines, including GeoCLIP~\citep{vivanco2024geoclip} for street view image geolocalization, RSVA~\citep{wang2022advancing} for infrastructure inference, RemoteCLIP~\citep{yeh2020using,remoteclip} for population prediction, LSTPM~\citep{sun2020go} for mobility prediction and MaxPressure~\citep{varaiya2013max}  for traffic signal control task.

\textbf{Evaluation Metrics}
We follow the common practice of each task to define the metrics. Metrics and instances for each task are presented in Figure~\ref{fig:category} and Table~\ref{table:tasks} in appendix. For each task with results from 13 cities, we report the mean value of them in Table~\ref{table:main:visual} and Table~\ref{table:main:text}. More detailed results like standard deviation value can be found in the appendix.

\begin{table}
\centering
\caption{Performance of 16 widely-used VLMs on four urban visual tasks in \textit{CityBench}. Here, `City' and `Loc.' represent the city name inference task and the geo-coordinates inference task for street view images, respectively; `Population' refers to the geospatial prediction task; `Infra' denotes the infrastructure inference task; and `Navigation' indicates the outdoor visual-language navigation task. `Succ.' stands for the success rate metric, while `Dist.' represents the distance metric.}
\label{table:main:visual}
\setlength{\tabcolsep}{0.5mm}
\resizebox{0.48\textwidth}{!}{
\begin{tabular}{lccccccc} 
\toprule
 \textbf{Tasks}& \multicolumn{4}{c}{\textbf{Perception\&Understanding}} &&\multicolumn{2}{c}{\textbf{Decision-making}}\\
 & \textbf{City}& \textbf{Loc.}& \multicolumn{2}{c}{\textbf{Population}}& \textbf{Infra}&\multicolumn{2}{c}{\textbf{ Navigation}}\\
 \textbf{Metrics}& \multicolumn{1}{c}{\textbf{Acc}$\uparrow$} & \textbf{Acc}$\uparrow$& \textbf{RMSE}$\downarrow$ &\textbf{$r^2$}$\uparrow$& \textbf{Acc}$\uparrow$&\textbf{Succ.}$\uparrow$ &\textbf{Dist.}$\downarrow$\\
 \midrule
 \rowcolor{lightgray}
 \textbf{Baselines}& & & & &  & & \\ 
 \textbf{GeoCLIP}& 0.340& 0.464& - &-& -  &- &- \\ 
 \textbf{RSVA}& - & - & -  &-& 0.655&- &- \\ 
 \textbf{RemoteCLIP}& - & - & \textbf{1.966} &\textbf{0.368}& -  &- &- \\ 
 \midrule
 \rowcolor{lightgray}
 \textbf{VLMs}& & & & &  & & \\
 \textbf{Qwen2VL-2B}& 0.630&0.407& 2.478 &0.008& 0.657&0.020 &679.333\\ 
 \textbf{InternVL2-2B}&0.238&0.380& 3.142 &-0.841& 0.738&0.247 &\uline{236.088}\\
 \textbf{InternVL2-4B}& 0.398& 0.397& 2.501 &-0.144& 0.735&\uline{0.260} &272.445\\
  \textbf{Yi-VL-6B}& 0.000& 0.105& 5.471 &-3.967& \uline{0.816}&\textbf{0.267} &429.683\\
  \textbf{Qwen2VL-7B}& 0.688& 0.522& 2.637 &-0.112&0.773&0.153 &529.549\\ 
 \textbf{LLaVANeXT-8B} & 0.267& 0.221& 3.31 &-0.764& 0.796&0.207 &361.647\\
  \textbf{MiniCPMV2.5-8B}  & 0.262& 0.223& 3.57 &-1.054& 0.806&\uline{0.260} &296.427\\
  \textbf{InternVL2-8B}& 0.522& \uline{0.728}& 2.806 &-0.320& 0.806&0.233 &\textbf{223.971}\\ 
  \textbf{GLM-4v-9B}& 0.726& 0.000& 2.769 &-0.516& \textbf{0.857}&0.247 &444.793\\
  \midrule
 \textbf{CogVLM2-19B}& 0.559& 0.326& 2.75 &-0.301& 0.726&0.087 &596.056\\
 \textbf{InternVL2-26B}&0.429&0.003& 2.683 &-0.209&0.790&0.180 &526.079\\ 
 \textbf{Yi-VL-34B}&0.251& 0.003&2.510 &-0.052&0.790&0.253 &384.005\\ 
 \textbf{LLaVANeXT-34B}& 0.501& 0.408& 2.61 &-0.163& 0.804&\textbf{0.267} &274.036\\
 \textbf{InternVL2-40B}& 0.574& 0.555&2.514 &-0.113& 0.808&0.213 &364.032\\ 
\midrule
 \textbf{Qwen-VL-plus} & \uline{0.793}& 0.645& 3.14 &-1.028& 0.454&0.240 &377.622\\
  \textbf{GPT4o}        & \textbf{0.862}& \textbf{0.797}& \uline{2.32} &\uline{0.122}& 0.812&0.180&388.582\\
\bottomrule
\end{tabular}}
\end{table}

\begin{table}
\centering
\caption{Performance of LLMs and VLMs on four urban tasks without visual input in \textit{CityBench}. Here, `Top 1' represents the Top-1 Accuracy metric, `Succ.' denotes the success rate metric, `Q' refers to the Queue Length metric, and `TP' indicates the throughput metric. `Und.' stands for the understanding task.}
\label{table:main:text}
\resizebox{0.48\textwidth}{!}{
\setlength{\tabcolsep}{0.5mm}
\begin{tabular}{llcccccccc} 
\toprule
 &\textbf{\textbf{Tasks}}& \textbf{Und.}& \multicolumn{6}{c}{\textbf{Planning \& Decision-making}}\\
 & & \textbf{GeoQA}& \multicolumn{2}{c}{\textbf{\uline{Mobility}}}& \multicolumn{2}{c}{\textbf{\uline{Exploration}}}&\multicolumn{2}{c}{\textbf{Traffic Signal}}\\
 &\textbf{Metrics}& \textbf{Acc}$\uparrow$& \textbf{Top1}$\uparrow$&\textbf{F1}$\uparrow$& \textbf{Succ.}$\uparrow$&\textbf{Steps}$\downarrow$& \textbf{Q}$\downarrow$& \textbf{TP}$\uparrow$\\
 \midrule
  \rowcolor{lightgray}
 & \textbf{Baselines} & & & & & & &\\
 &\textbf{LSTPM}& -& 0.114 & 0.086 & - &-&- & -\\
 &\textbf{Fixed-Time}& -& - &- & - &-&57.870 &993.333 \\
 &\textbf{Max-Pressure}& -& - &-& - &-&\textbf{36.898} &\textbf{1345.333}\\
\midrule
 \rowcolor{lightgray}
& \textbf{LLMs} & & & & & & &\\
&\textbf{Mistral-7B} & 0.229& 0.090 & 0.087 & 0.730 & 5.382 & 64.120 &853.333\\
&\textbf{Qwen2-7B} &  0.289& 0.142 & 0.109 &  0.697 & 5.889 & 62.271 &880.000\\
 & \textbf{Intern2.5-7B} &  0.304& 0.118 & 0.102 &  0.738 & 5.552 & 55.121 & 1047.667\\
 &\textbf{LLama3-8B} & 0.297& 0.130 & 0.094 & 0.747 & 5.304 & 57.738 &1014.333\\
 & \textbf{Gemma2-9B} &  0.339& 0.131 & 0.120 &  0.716 & 5.679 & 74.475 &651.333\\
  \cmidrule{2-9}
 &\textbf{Intern2.5-20B} &  0.315& 0.116 & 0.098 &  0.679 & 6.243 & 61.229 &958.667\\
  &\textbf{Gemma2-27B} &  0.349& 0.145 & 0.118 &  0.713 & 5.733 & 56.081 &1009.333\\
 & \textbf{Qwen2-72B} &  0.357& 0.155 & \uline{0.135} &  0.697 & 5.887 & 66.924 &793.333\\
 & \textbf{LLama3-70B} & 0.329& \textbf{0.159} & 0.130 & \textbf{0.796} & \textbf{4.941}& 59.338 &959.667\\
  &\textbf{Mixtral-8x22B} & 0.321& \uline{0.155} & \textbf{0.136} & 0.745 & 5.339 & 65.682 &821.333\\ 
 &\textbf{DeepSeekV2} & \uline{0.358}& 0.126 & 0.101 & 0.698 & 5.739 & 56.086 &1020.333\\
 \midrule
 \rowcolor{lightgray}
 & \textbf{VLMs} & & & & & & & \\
 &\textbf{InternVL2-2B} & 0.296 & 0.000 & 0.000 & 0.672 & 6.015 & 55.725 & 1012.000 \\ 
 &\textbf{InternVL2-4B} & 0.304 & 0.130 & 0.102 & 0.674 & 6.091 & 74.499 & 647.667 \\ 
 &\textbf{InternVL2-8B} & 0.329 & 0.142 & 0.102 & 0.703 & 5.714 & 53.196 & 1069.667 \\ 
 &\textbf{InternVL2-26B} & 0.310 & 0.137 & 0.107 & 0.694 & 5.723 & 57.512 & 971.667 \\ 
 &\textbf{InternVL2-40B} & 0.351 & 0.159 & 0.121 & 0.675 & 6.041 & \uline{52.459} & \uline{1087.000} \\ 
 & \textbf{Qwen2VL-2B} & 0.293 & 0.103 & 0.075 & 0.643 & 6.315 & 56.097 & 1003.667 \\ 
 &\textbf{Qwen2VL-7B} & 0.286 & 0.144 & 0.102 & 0.660 & 6.155 & 55.885 & 995.333 \\
 &\textbf{MiniCPMV2.5-8B} & 0.308 & 0.124 & 0.092 & 0.708 & 5.643 & 56.066 & 1001.000 \\
 &\textbf{LLaVANeXT-8B} & 0.313 & 0.124 & 0.084 & 0.688 & 5.891 & 56.184 & 989.333 \\
 &\textbf{GLM-4v-9B} & 0.296 & 0.133 & 0.092 & 0.680 & 5.979 & 53.870 & 1058.000 \\
 &\textbf{CogVLM2-19B} & 0.282 & 0.026 & 0.029 & 0.710 & 5.905 & 55.229 & 1046.667 \\
 \midrule
 &\textbf{GPT3.5-Turbo} & 0.285& 0.152 & 0.113 & 0.719 & 5.473 & 56.219 &1022.000\\
 & \textbf{GPT4-Turbo} & \textbf{0.398}& 0.147 & 0.125& \uline{0.757} & \uline{5.184} & 55.761 &1022.333\\
\bottomrule
\end{tabular}}
\end{table}

\subsection{Overall Performance on CityBench}

\subsubsection{\textbf{CityBench are Challenging for LLMs and VLMs}}
The performance of LLMs and VLMs on \textit{CityBench} is summarized in Table~\ref{table:main:text} for urban tasks without visual input and in Table~\ref{table:main:visual} for urban visual tasks. As we can observe, except for the street view image geolocalization and infrastructure inference tasks, the performance of LLMs on the remaining six tasks is suboptimal and remains far from the ideal ceiling. For instance, the performance on GeoQA task, which evaluates detailed knowledge about urban elements of LLMs, is only 0.398, significantly lower than the best possible score of 1.0. These results demonstrate that CityBench is a challenging benchmark on urban tasks for LLMs.

\subsubsection{\textbf{LLMs and VLMs Struggle with Numerical Tasks}} As shown in Table~\ref{table:main:text} and Table~\ref{table:main:visual}, the performance of LLMs and VLMs on numerical tasks, including population estimation and traffic signal control, significantly lags behind existing baselines. In population estimation tasks, the best-performing LLM, GPT-4o, underperforms RemoteCLIP by 18\% in terms of RMSE. Similarly, in traffic signal control tasks, the top-performing VLM, InternVL2-40B, trails behind the Max-Pressure method by 41.9\% in queue length. Therefore, improving the performance of LLMs on numerical tasks is crucial for their application in urban tasks. Although some studies have explored potential solutions by adding task-specific heads to LLMs, such designs may compromise the generalizability of the model.

\subsubsection{\textbf{Performance Consistency between Various Urban Tasks}}
As the best-performing VLM in urban visual tasks, GPT-4o only excels in the first two tasks, while other VLMs, such as GLM-4v-9B, LLaVANeXT-34B, and InternVL2-8B, outperform it in the remaining two tasks. For urban tasks without visual input, LLama3-70B achieves the best performance in mobility prediction and urban exploration tasks, but it falls behind other high-performing LLMs like GPT-4 Turbo and InternVL2-40B in the other two tasks. In other words, due to the heterogeneity and complexity of urban tasks, no LLM or VLM, including the powerful GPT-4 series, can consistently perform well across all tasks. These results highlight the challenges posed by CityBench and underscore the necessity of developing domain-specific LLMs and VLMs tailored for urban tasks.

\subsubsection{\textbf{LLMs and VLMs Exhibit Geospatial Bias}} \label{sec:bias}
\begin{figure}
    \centering
    \begin{subfigure}{.48\textwidth}
        \includegraphics[width=\textwidth]{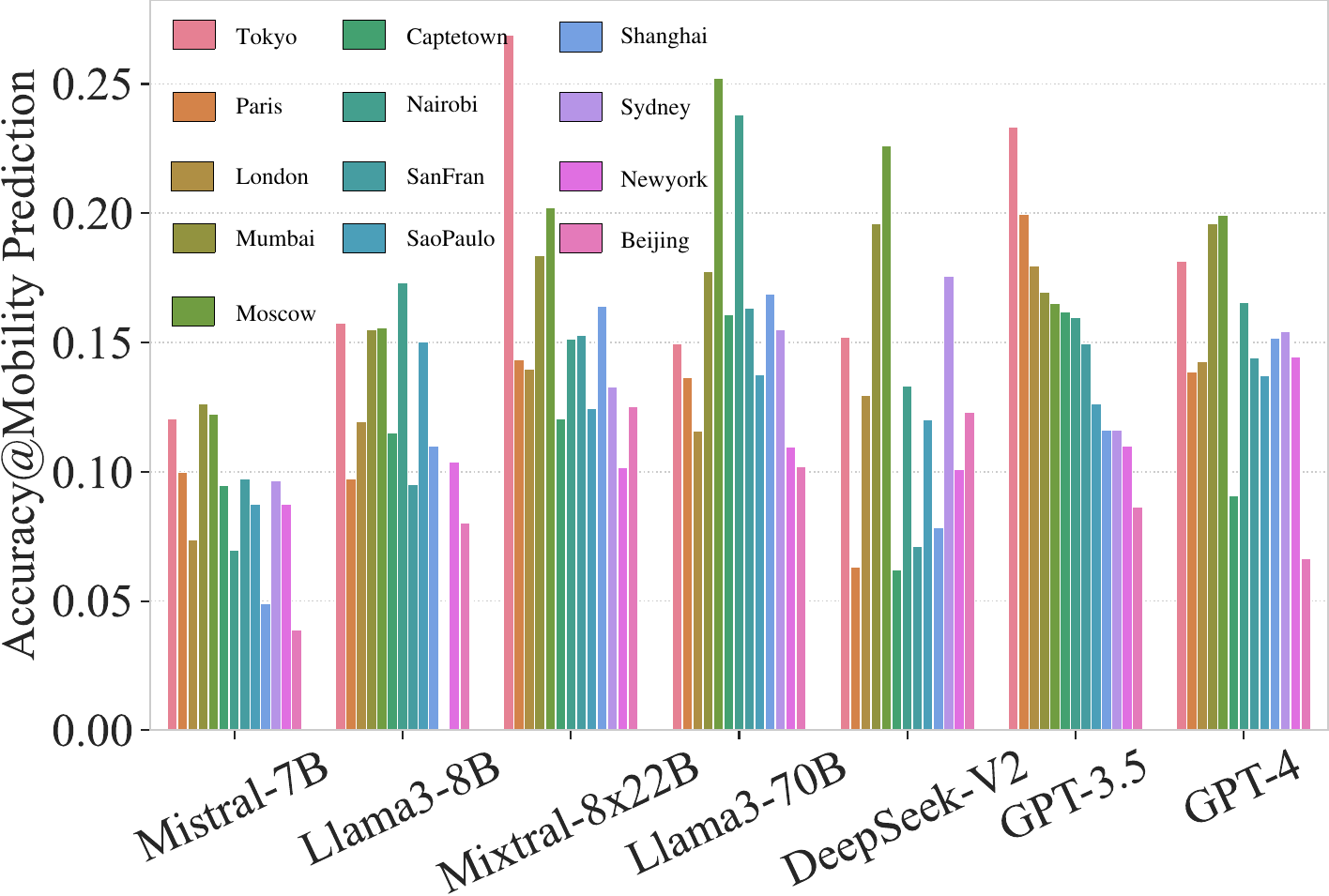}
    \end{subfigure}
    \hfill
    \begin{subfigure}{.48\textwidth}
        \includegraphics[width=\textwidth]{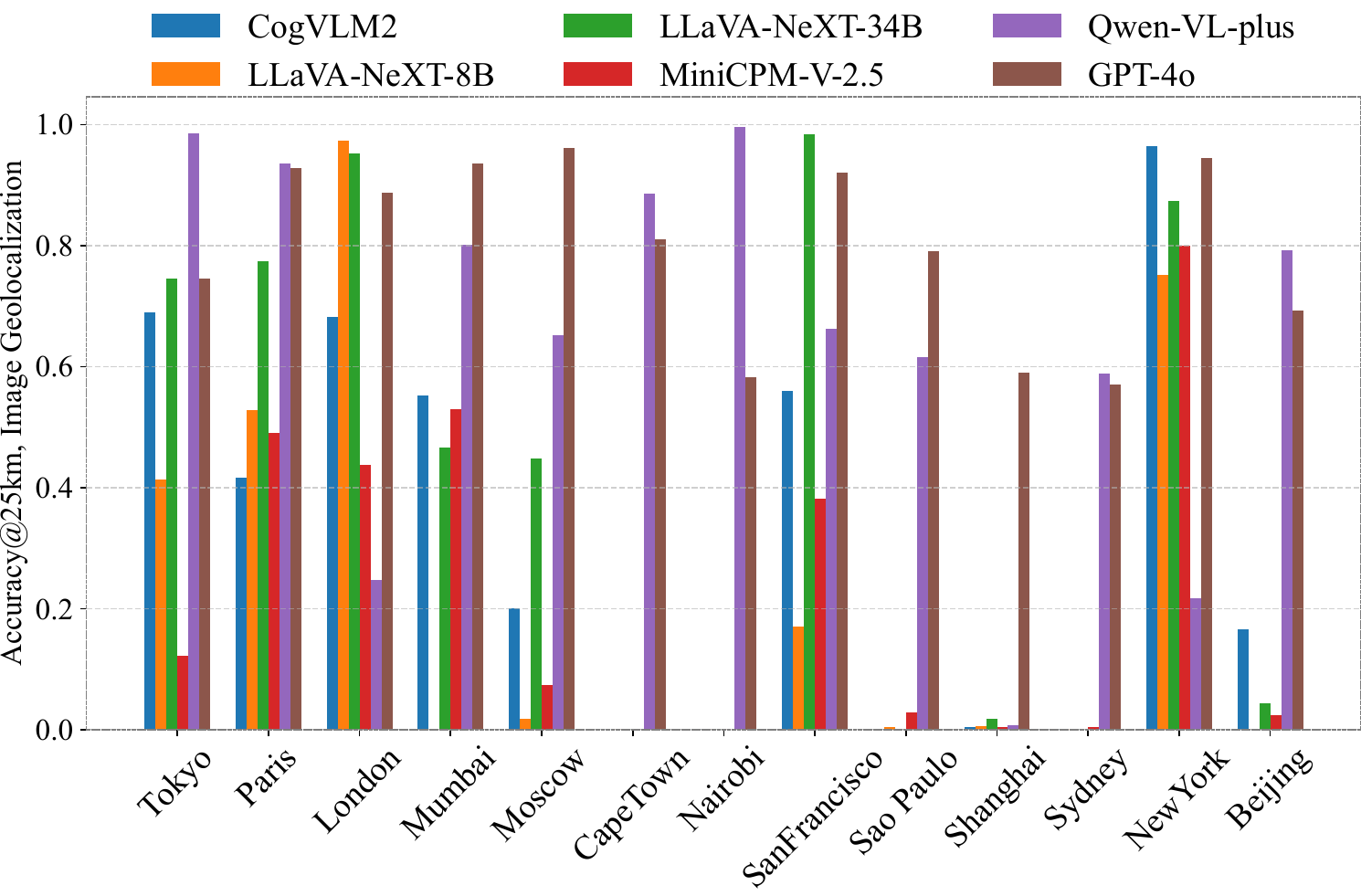}
    \end{subfigure}

    \caption{Detailed performance results of LLMs on two tasks: (top) mobility prediction and (bottom) image geolocalization. Both tasks are evaluated across multiple cities and multiple models, demonstrating that significant performance variations across diverse urban contexts are consistently observed even with different model architectures, highlighting the pervasive nature of geospatial bias in these models.}
    \label{fig:ablation_study}
\end{figure}

To further investigate the difference between LLMs and VLMs, we report the detailed results of mobility prediction task and image geolocalization tasks from 13 cities in Figure~\ref{fig:ablation_study}. Based on the above results, we have made several interesting discoveries. First, we find that the performance of different LLMs varies a lot across different cities, no LLM can always perform best in mobility prediction tasks. Second, we find that the performance of VLMs on visual task like image geolocalization task are significantly biased.  Most VLMs perform well in major international cities, but poorly in some lesser-known cities (e.g., CapeTown and Nairobi). We provide preliminary evidence for this phenomenon by analyzing the number of publicly accessible websites indexed on Google and Wikipedia. We find that cities with fewer publicly available websites tend to pose greater challenges for LLMs, while those with more extensive online presence are generally easier for the models to handle. For more details, see section~\ref{sec:bias:google} in the appendix.  The variability in evaluation results demonstrate the necessity of establishing a global evaluation benchmark, and also highlights the potential shortcomings and areas for improvement of LLMs.

\subsection{Detailed Analysis between Models}

\subsubsection{\textbf{Disparity between Proprietary and Open-source Models}}
Although the GPT-4 series performs well on most tasks, several open-source models have achieved better results on multiple other tasks, albeit not consistently by a single open-source model. In other words, we do not observe a dominant advantage of proprietary models over open-source models on \textit{CityBench}, which may be attributed to the uniqueness and heterogeneity of urban tasks. The reasons behind this phenomenon warrant further in-depth analysis in future research. 
While the performance of models within the same series generally follows the scaling law with respect to model size, most VLMs' performance across tasks is not robust. For instance, some VLMs exhibit near-zero performance, and larger models within the same series do not always outperform smaller ones. For example, InternVL2-26B performs worse than InternVL2-8B. Similar trends are observed in LLMs, where Intern2.5-20B does not consistently outperform Intern2.5-7B.

\subsubsection{\textbf{Performance Correlation between VLMs and LLMs}}
While most VLMs are continuously trained based on LLMs, we investigate the impact of LLMs on VLM performance. First, we find that the performance variability across different LLMs and VLMs is primarily influenced by the capabilities of the LLM backbone. For instance, in widely used VLMs for urban tasks, Intern2.5-7B consistently outperforms Qwen2-7B and Mistral-7B across most tasks, which can be attributed to Intern2.5-7B's superior performance in general NLP tasks at the same parameter scale. Similarly, LLaVA-NeXT-8B demonstrates performance comparable to MiniCPM-V2.5-8B, as both models share the same LLM backbone, LLaMA3-8B. Second, as shown in Table~\ref{table:main:text}, we observe that most VLMs underperform compared to their LLM bases on urban tasks without visual input. This demonstrates that while post-training of LLM-based VLMs enhances visual capabilities, it inevitably leads to performance degradation in original textual tasks. For example, LLaVA-NeXT-8B lags behind LLaMA3-8B on all textual tasks, with an average performance degradation of 4.10\%, while MiniCPMv2.5-8B exhibits a smaller degradation of over 1.88\%. Therefore, maintaining the general capabilities of LLMs during the training of VLMs should be a key direction to ensure their effectiveness across a wide range of tasks.

\begin{figure*}[htbp]
    \centering
    \includegraphics[width=1.\textwidth]{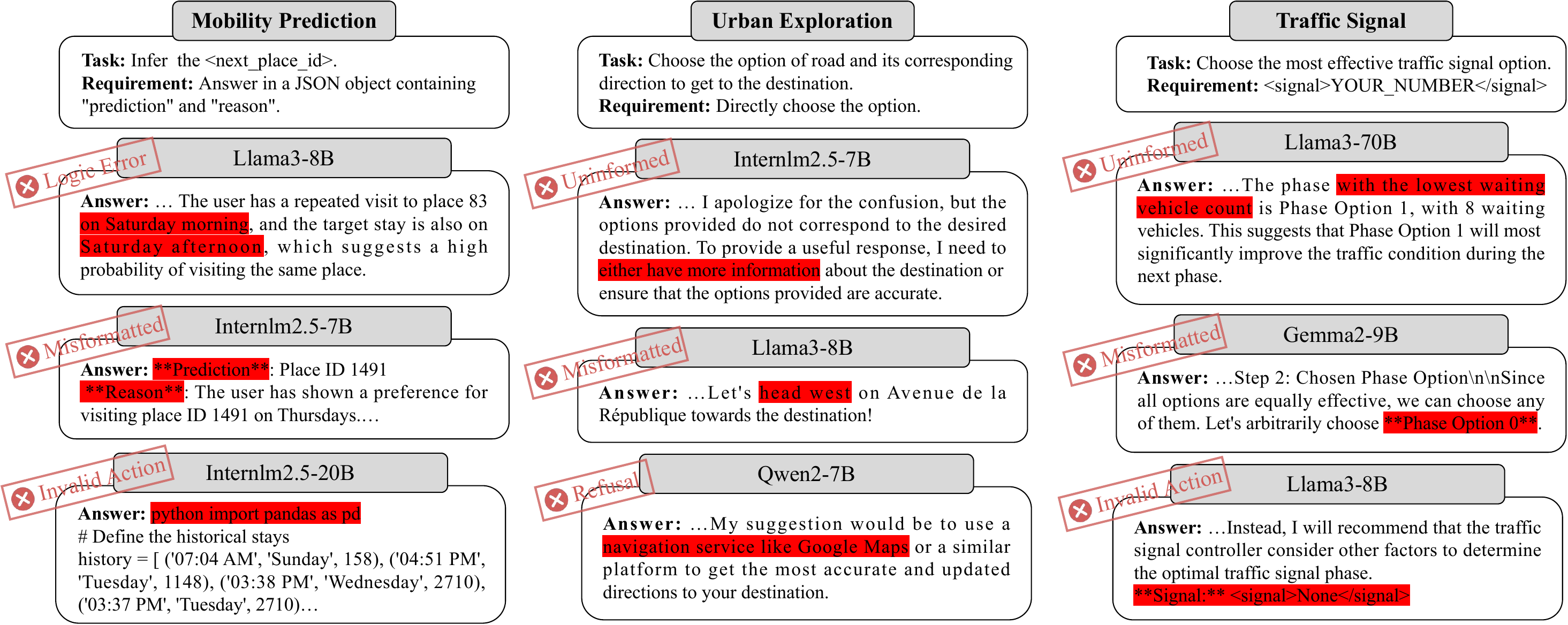}
    \caption{Error analysis in mobility prediction, urban exploration, and traffic signal control tasks reveals common issues: logic errors, format errors, invalid actions, refusal to answer, and hallucinations. Full prompts for each task are in the appendix.
    }
    \label{fig:error}
\end{figure*}

\subsubsection{\textbf{Typical Errors of LLMs and VLMs in CityBench}} \label{sec:error}
We find that LLMs often display errors such as \textit{logic error}, \textit{format error}, \textit{invalid action}, \textit{refusal to answer}, and \textit{hallucinations}. The types of errors are highly correlated with the characteristics of the LLMs. For instance, certain models, such as MiniCPM2.5-8B, exhibit excessive alignment in handling geospatial-related content, leading to a systematic refusal to respond to queries across various tasks, as illustrated in the image localization task depicted in Figure~\ref{fig:ablation_study}. On the other hand, smaller models like InternVL2-2B often struggle to follow instructions, leading to format errors and invalid actions. We present typical error cases in Figure~\ref{fig:error}. For example,  Llama3-8B exhibited a logical error in its judgment of time in the task mobility prediction. For the urban exploration task, Qwen2-7B refused to choose the option and instead demanded the user to use a navigation service to solve the problem. Intern2.5-7B directly stated that it lacks expertise in this area and needs more information to answer the question. Llama3-8B provided an invalid option in the traffic signal task, rendering \textit{CitySimu} unable to perform next action. We notice that the most frequent error is Misformatted, and several instances close to 0 in Table~\ref{table:main:text} are mostly caused by formatting errors. Thus, one of the promising direction is to reduce these error from LLMs to improve their practicality. More detailed analysis on VLMs are presented in section~\ref{sec:err-vlm} of appendix.

\section{Related Work}\label{sec:realted}

\textbf{Evaluating LLMs for Urban Knowledge and Tasks.} Researchers from various urban related fields have conducted extensive evaluations of LLM in urban space from different aspects~\citep{feng2025survey,ding2024understanding}. Kuckreja et al.~\citep{kuckreja2023geochat} evaluate the performance of multi-modal LLMs on several remote sensing related tasks. Yang et al.~\citep{yang2024v} propose V-IRL benchmark to evaluate the performance of multi-modal LLMs on street view image related tasks including localization and recognition tasks. Mai et al.~\citep{mai2023opportunities} and Manvi et al.~\citep{manvi2023geollm} use LLMs to predict social indicators like population and education level. Gurnee et al.~\citep{gurnee2023language} and Bhandari et al.~\citep{bhandari2023large} try to testify whether LLMs know the coordinates of geospatial entity. Mooney et al.~\citep{mooney2023towards} and Deng et al.~\citep{deng2023learning} use GIS exams to understand the geospatial skills of LLMs. 
Different from these works, we first introduce the interactive simulator based systematic evaluation system for LLMs and VLMs, which covers various data modalities, diverse urban task types and differentiated data from 13 cities around the world. 

\noindent
\textbf{Interactive Decision-making and Urban Simulator.} Beyond the above static evaluation, researchers also evaluate the capacity of LLMs in the interactive decision making tasks with customized simulators, e.g., web agent~\citep{liu2023agentbench} with web environment and embodied intelligence~\citep{yang2023learning} with virtual home~\citep{puig2018virtualhome}. In the urban domain, Schumann et al.~\citep{schumann2024velma} apply LLM to do the visual language navigation task in Touchdown~\citep{chen2019touchdown} and Lai~\citep{lai2023large} apply LLMs as the traffic light controller in CityFlow~\citep{zhang2019cityflow} to manage the road traffic. Besides, Yang et al.~\citep{yang2024v} design V-IRL as the environment of street view image related tasks and propose a global scale virtual intelligence benchmark. These works only evaluate the potential of LLMs in single urban decision-making task and most of their results rely on small-scale datasets in limited regions. Different from them, our work builds on an efficient urban simulator with global scale and supports 4 representative urban decision-making tasks with different modality in one benchmark, including urban exploration, outdoor navigation, mobility prediction and traffic control task.

\section{Conclusion}
In this paper, we propose \textit{CityBench}, a systematic evaluation benchmark for LLMs and VLMs in diverse urban tasks. With the data support from \textit{CityData} and simulation support from \textit{CitySimu}, we design 8 important urban tasks in 13 cities to constitute the \textit{CityBench} for evaluating the capabilities of LLMs and VLMs. Extensive experiments present that LLMs and VLMs exhibit exceptional performance in various urban tasks requiring commonsense and semantic understanding, but fail in challenging urban tasks which require professional domain knowledge and precise numeric calculations. The extensive results from \textit{CityBench} demonstrate the potential the applying LLMs and VLMs in various urban tasks and also shed light for the future research of developing more powerful LLMs and VLMs for urban tasks.

\bibliographystyle{ACM-Reference-Format}
\balance
\bibliography{references}

\clearpage
\appendix
\section{Appendix} \label{sec:app}

\subsection{ Few-Shot Performance of LLMs in CityBench}
Here, we present the few-shot performance of several representative LLMs in Beijing in the Table~\ref{table:appendix:fewshot}. For all text-based tasks, we use 2-shot as the default few-shot method. As shown in the table, the impact of few-shot learning varies across different models and tasks. For instance, few-shot learning improves performance for Gemma-27B in GeoQA but reduces it for Gemma2-9B on the same task. Similarly, it benefits Gemma2 in traffic signal tasks but proves detrimental for Llama3.

\begin{table*}[!htbp]
\centering
\caption{ Few-shot performance of several representative LLMs in Beijing.}
\label{table:appendix:fewshot}
\resizebox{0.9\textwidth}{!}{
\begin{tabular}{ccccccccc} 
\toprule
\multirow{2}{*}{\textbf{Model@Beijing}} & \textbf{GeoQA} & \multicolumn{2}{c}{\textbf{Mobility Prediction}} & \multicolumn{2}{c}{\textbf{Urban Exploration}} & \multicolumn{3}{c}{\textbf{Traffic Signal}} \\ 
\cmidrule{2-9}
 & \textbf{Accuracy} & \textbf{Top1 Acc} & \textbf{F1} & \textbf{Succ Rate} & \textbf{Steps} & \textbf{Throughput} & \textbf{Travel Time} & \textbf{Queue Length} \\ 
\midrule
\textbf{Gemma2-9B-fewshot} & 0.306& 0.115& 0.101& 0.728 & 6.152 & 1528 & 2129.21 & 64.269 \\ 
\textbf{Gemma2-9B-zeroshot} & 0.339 & 0.131 & 0.120 & 0.716 & 5.679 & 1448 & 2214.451 & 71.729 \\
\midrule
\textbf{Gemma2-27B-fewshot} & 0.359& 0.109& 0.076& 0.696 & 6.260& 2240 & 1746.89 & 31.683 \\ 
\textbf{Gemma2-27B-zeroshot} & 0.349 & 0.145 & 0.118 & 0.713 & 5.733 & 2187 & 1762.182 & 33.322 \\ 
\midrule
\textbf{LLama3-8B-fewshot} & 0.288& 0.095& 0.0546 & 0.692 & 6.424 & 1547 & 2132.233 & 61.84 \\ 
\textbf{LLama3-8B-zeroshot} & 0.297 & 0.130 & 0.094 & 0.747 & 5.304 & 2128 & 1873.757 & 40.941 \\ 
\midrule
\textbf{Llama3-70B-fewshot} & 0.343& 0.089& 0.0620 & 0.796 & 4.941 & 1810 & 1962.493 & 50.541 \\ 
\textbf{Llama3-70B-zeroshot} & 0.329 & 0.159 & 0.130 & 0.74 & 5.876 & 2031 & 1893.517 & 43.475 \\
\bottomrule
\end{tabular}}
\end{table*}

\subsection{Details of Quality Control}
In CityBench, the authors will participate in the quality control process for some tasks, following the automatic quality control stage. Taking the manual checking of GeoQA as an example, the original data from OpenStreetMap contains low-quality information, with missing or incorrect details about AOI, POI, and roads. When this low-quality data is used in the evaluation task, LLMs may become confused and generate meaningless answers. In such cases, the authors review the questions to ensure that the information in the context is meaningful. However, due to time limitations for participants, we can only randomly sample the evaluation cases. For instances from cities and regions where authors are unfamiliar, we filter out low-quality instances. For instances from cities and regions where authors are familiar, we rewrite low-quality instances using external information, such as commercial map services.  Finally, if data quality remains unsatisfactory after filtering and rewriting, we will regenerate a certain number of cases to fill the gaps. In fact, while considering the probabilities of filtering, we generate enough candidate instances during the initial generation.

\subsection{Additional Results of Geospatial Bias Analysis} \label{sec:bias:google}
Taking three well-performing cities (New York, London, Paris) and three under-performing cities (Shanghai, CapeTown, Nairobi) as examples, Table~\ref{tab:google} illustrates the relationship between the performance of street view image localization tasks and the size of the training corpus for LLMs. The training corpus size is approximated by the number of Google search entries and Wikipedia entries for each city. Compared to the well-performing cities, the under-performing cities have significantly smaller `training corpora' in the public websites. Additionally, open-source models exhibit significantly worse performance than commercial models in terms of geospatial bias. This observation provides initial evidence for analyzing geographical bias, but we believe there are more diverse factors contributing to this phenomenon.

\begin{table*}[!htbp]
\centering

\caption{ The performance of street view image localization tasks in different cities and its relationship with the amount of data in the `training corpus' of LLMs, where the amount of data in the training corpus is approximated by the number of Google search entries and Wikipedia entries for each city.}
\label{tab:google}
\resizebox{0.6\textwidth}{!}{
\begin{tabular}{lrrrr} 
\toprule
\textbf{Cities} & \textbf{GPT4o} & \textbf{MiniCPM-V2.5} & \textbf{Google Search Pages} & \textbf{Wikipedia Pages} \\ 
\midrule
\textbf{Shanghai} &  59\%&  0.4\%& 582,000,000 & 62,495 \\
\textbf{CapeTown} &  81\%&  0& 587,000,000 & 54,724 \\
\textbf{Nairobi} &  58.19\%&  0& 205,000,000 & 14,226 \\
\textbf{New York} &  94.39\% &  80\%& 7,070,000,000 & 1,039,264 \\
\textbf{London} &  88.80\%&  43.8\%& 5,460,000,000 & 848,381 \\
\textbf{Paris} &  92.80\%&  49\%& 4,170,000,000 & 353,500 \\
\bottomrule
\end{tabular}}
\end{table*}

\subsection{Error Analysis of VLMs} \label{sec:err-vlm}
\begin{figure*}[!htbp]
    \centering
    \includegraphics[width=0.9\textwidth]{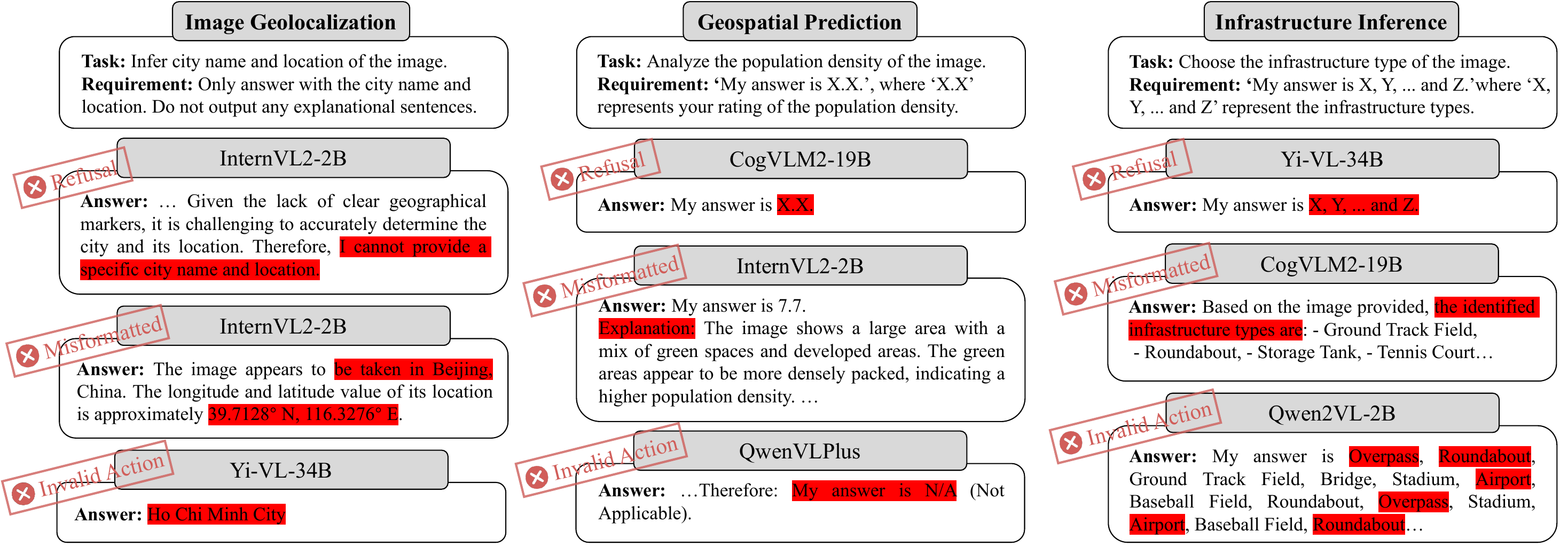}
    \caption{ Error analysis in image geolocalization, geospatial prediction and infrastructure inference tasks. 
    }
    \label{fig:error-vlm}
\end{figure*}

Urban visual tasks require the model to make decisions directly without going through an explanation process. As a result, as shown in Figure~\ref{fig:error-vlm}, common errors include format errors, invalid actions, refusal to answer and hallucinations. In the image geolocalization task, InternVL2-2B provides a response that do not follow the required format, while Yi-VL-34B gives an irrelevant invalid response. CogVLM2-19B and Yi-VL-34B, in the geospatial prediction and infrastructure inference tasks respectively, repeat the examples provided in the question and refuse to answer the actual question. Due to the response format requirements of the tasks, the most common error made by VLMs in urban visual tasks is misformatted responses.

\subsection{Map building tool in \textit{CityData}}
The map building tool~\footnote{\url{https://github.com/tsinghua-fib-lab/mosstool}} enhances open-source map data to support subsequent behavior simulations, encompassing lane topology recovery, relationship recognition, intersection reconstruction, area of interest (AOI) mapping, point of interest (POI) clustering, basic traffic rule generation, and right-of-way construction.

\begin{table*}[!htbp]
\centering

\caption{{Detailed information of 8 evaluation tasks in \textit{CityBench}, including data modality, metric and data instances. Task settings across different cities keep consistent.}}
\label{table:tasks}
\resizebox{0.75\textwidth}{!}{
\setlength{\tabcolsep}{0.1mm}
\begin{tabular}{cccccc} 
\toprule
\textbf{CityBench} & \textbf{Tasks} & \textbf{Modality} & \textbf{Metrics} & \textbf{Instances}  &\textbf{Images}\\
\midrule
\multirow{4}{*}{\begin{tabular}[c]{@{}l@{}}\textbf{Perception\&}\\\textbf{Understanding }\end{tabular}} 
 & Image Geolocalization & Image & Acc, Acc@1km/25km & 6500&6500\\
 & Geospatial Prediction & Image & $r^2$, RMSE & 5739
&5739
\\
 & Infrastructure Inference & Image & Accuracy, Recall & 5739&5739
\\
 & GeoQA for City Elements & Text & Accuracy & 13126&/\\
 \midrule
\multirow{4}{*}{\begin{tabular}[c]{@{}l@{}}\textbf{Planning\&}\\\textbf{Decision Making }\end{tabular}} 
 & Mobility Prediction & Text & Top1-Acc, F1 & 6500&/
\\
 & Urban Exploration & Text & Steps, Success Rate & 650
&/ \\
 & Outdoor Navigation & Image & Distance, Steps, Success Rate & 650&55984
\\
 & Traffic Signal Control & Text & Queue Length, Throughput & 1hour $\times$ 13 cities &/
\\
\bottomrule
\end{tabular}}
\end{table*}

\subsection{Discussion} \label{sec:diss}
We discuss some limitation of current work as below.

\noindent
\textbf{Limitations.} While our platform is based on the public data from various sources, the quality of different data may play a important role in the evaluation results. In the future, we plan to collect more kinds of tasks with global scale groundtruth data to further improve the reliability and representativeness of benchmark.

\noindent
\textbf{Ethical considerations and potential societal impact.} Our benchmark is designed for enable the global evaluation of LLMs and VLMs for various cities with different cultures and countries. We try our best to improve the ease-of-use and fairness for cities with different development levels. However, due to the limitation of accessed data, the evaluation results for different cites varies a lot. Therefore, the variation in evaluation results caused by data quality may lead to a certain degree of misunderstanding regarding the performance on some urban problems. We call the whole community for attention to this issue to improve the usability of LLMs across different races and countries, promoting fairness and sustainable development of the world.

\noindent
\textbf{Develop foundation model for urban domain.} Based on the results of our benchmark, we find existing LLMs perform poorly on many urban tasks, even worse than some classic simple baseline algorithms. Developing LLMs tailored for urban domain is urgently necessary. We hope our benchmark can accelerate this development and we look forward to a more comprehensive and robust evaluation framework for urban domain.

\end{document}